\documentclass[conference]{IEEEtran}
\IEEEoverridecommandlockouts
\usepackage{cite}
\usepackage{amsmath,amssymb,amsfonts}
\usepackage{graphicx}
\usepackage[caption=false]{subfig}
\usepackage[utf8]{inputenc}
\usepackage[T1]{fontenc}
\usepackage{bbm}

\DeclareMathOperator*{\argmin}{arg\,min}

\usepackage{booktabs}
\usepackage{xcolor}
\usepackage{tikz}
\usepackage{float}
\usepackage{textcomp}
\usepackage{tikz}
\usepackage{textcomp}
\usepackage{scalerel}
\usepackage[ruled,linesnumbered]{algorithm2e}
\usepackage{algpseudocode}
\usepackage{array}
\usepackage{graphicx}
\usepackage{bbding}
\usepackage{mwe}
\usepackage{caption}
\usepackage{stfloats} 
\usepackage{lipsum}
\usepackage{graphicx}
\usepackage{kantlipsum}
\usepackage{multirow}	        
\usetikzlibrary{svg.path}

\definecolor{orcidlogocol}{HTML}{A6CE39}
\tikzset{
  orcidlogo/.pic={
    \fill[orcidlogocol] svg{M256,128c0,70.7-57.3,128-128,128C57.3,256,0,198.7,0,128C0,57.3,57.3,0,128,0C198.7,0,256,57.3,256,128z};
    \fill[white] svg{M86.3,186.2H70.9V79.1h15.4v48.4V186.2z}
                 svg{M108.9,79.1h41.6c39.6,0,57,28.3,57,53.6c0,27.5-21.5,53.6-56.8,53.6h-41.8V79.1z M124.3,172.4h24.5c34.9,0,42.9-26.5,42.9-39.7c0-21.5-13.7-39.7-43.7-39.7h-23.7V172.4z} svg{M88.7,56.8c0,5.5-4.5,10.1-10.1,10.1c-5.6,0-10.1-4.6-10.1-10.1c0-5.6,4.5-10.1,10.1-10.1C84.2,46.7,88.7,51.3,88.7,56.8z};
  }
}
\newcommand\orcidicon[1]{\href{https://orcid.org/#1}{\mbox{\scalerel*{
\begin{tikzpicture}[yscale=-1,transform shape]
\pic{orcidlogo};
\end{tikzpicture}
}{|}}}}

\usepackage{hyperref} 

\begin{document}
\title{Dynamic Clustering in Federated Learning
}
\author{\IEEEauthorblockN{
 Yeongwoo Kim\IEEEauthorrefmark{1}\IEEEauthorrefmark{2}
, Ezeddin Al Hakim\IEEEauthorrefmark{2}, Johan Haraldson\IEEEauthorrefmark{2}, Henrik Eriksson\IEEEauthorrefmark{2},\\
José Mairton B. da Silva Jr.\IEEEauthorrefmark{1}, Carlo Fischione\IEEEauthorrefmark{1},\\
\IEEEauthorrefmark{1}KTH Royal Institute of Technology, Stockholm, Sweden\\
\IEEEauthorrefmark{2}Ericsson Research, Stockholm, Sweden\\}
}

\maketitle
\begingroup\renewcommand\thefootnote{\textsection}
\thispagestyle{empty}
\pagestyle{empty}

\begin{abstract}
In the resource management of wireless networks, Federated Learning has been used to predict handovers. However, non-independent and identically distributed data degrade the accuracy performance of such predictions. To overcome the problem, Federated Learning can leverage data clustering algorithms and build a machine learning model for each cluster. However, traditional data clustering algorithms, when applied to the handover prediction, exhibit three main limitations: the risk of data privacy breach, the fixed shape of clusters, and the non-adaptive number of clusters. To overcome these limitations, in this paper, we propose a three-phased data clustering algorithm, namely: generative adversarial network-based clustering, cluster calibration, and cluster division. We show that the generative adversarial network-based clustering preserves privacy. The cluster calibration deals with dynamic environments by modifying clusters. Moreover, the divisive clustering explores the different number of clusters by repeatedly selecting and dividing a cluster into multiple clusters. A baseline algorithm and our algorithm are tested on a time series forecasting task. We show that our algorithm improves the performance of forecasting models, including cellular network handover, by 43\%.

\end{abstract}
\vspace{-3mm}
\section{Introduction}

\label{sec:intro}

Machine learning (ML) is emerging as a key theory for the resource management in wireless networks. One of the most important functions for resource management is the ability to forecast the load of network resources. This can be done by using the time series from the wireless networks (e.g., 5G or cellular networks) where the varying load is recorded in the form of the time series.
In fact, there have been many attempts to forecast and manage the load of network resources at access point or base station (BS) to ensure a high quality of service \cite{hussain2020machine}. However, current load forecasting methods use model-based approaches and have shown low prediction accuracy \cite{boutaba2018comprehensive}. This motivates to explore the use of ML algorithms for network load forecasting, as an alternative to model based approaches. 

Due to the limitations of centralized ML algorithms and the dynamic characteristics of wireless networks \cite{konevcny2015federated}, Federated learning (FL) is a ML method that is gaining popularity. In centralized ML, there is a unit called server that performs the computations, and distributed units called clients that send data to the server. Such an approach is problematic with data from BSs. In fact, when BSs are clients and one of the BSs is selected as server, centralized ML algorithms require data curation from clients to a server, which causes the risk of client's data privacy breach. There is also an inference delay one would need to consider when ML decisions are taken centrally. These aspects have motivated the development of FL which trains and saves ML models on clients. Thus, the need to transmit the data (e.g, the load of the network resources) from clients to a server is reduced, and the decisions are made on clients. 

However, non-independent and identically distributed (non-IID) data degrades the performance of the ML models in FL \cite{zhao2018federated}. The degradation must be addressed since we cannot expect an IID distribution in real-world wireless networks. To be specific, the characteristics of data in each BS can be different across BSs. To deal with such situation, an approach consists in grouping the data of different BSs into clusters and in training a model per each group, as it has been investigated in these works: \cite{huang2019patient, briggs2020federated, ghosh2019robust, mansour2020three}. Although these works are promising, they also present the following important limitations: 
\begin{itemize}
    \item \textbf{Non-adaptive number of clusters}: Traditional clustering algorithms create a fixed number of clusters;
    \item \textbf{Risk of data privacy breach}: The data can be eavesdropped using the data transmitted over a network;  
    \item \textbf{Fixed shapes of clusters}: The data generated by the clients is generally time varying and their characteristics, as used in traditional clustering algorithms, change over time. Thus, in such dynamic situations, the fixed size of data clusters is clearly sub-optimal.
\end{itemize} 

This paper proposes to address the three aforementioned limitations. Our main contribution is a novel clustering framework which preserves data privacy, creates dynamic clusters, and adapts the number of clusters over time. The framework consists of the following phases:

\begin{itemize}
    \item \textbf{Phase 1}: generative adversarial network (GAN)-based clustering in \cite{mukherjee2019clustergan} forms clusters without sharing raw data; 
    \item \textbf{Phase 2}: Cluster calibration in \cite{mansour2020three} dynamically relocates clients to update clusters;
    \item \textbf{Phase 3}: Cluster division in \cite{savaresi2002cluster} selects a cluster and divides it into multiple clusters.
\end{itemize} 

To validate our algorithm, we compare it to the baseline clustering algorithm for time series in \cite{hyndman2015large}. From the comparison, we show the improved network handover forecasting (average improvement of 43\%), and other benchmark use cases such as  the demand for electrical power, the trace of pens, and the number of pedestrians.

The remainder of this paper is organized as follows. Section~\ref{sec:related_work} overviews previous works for non-IID data and clustering in FL. In Section~\ref{sec:background}, we introduce studies from the concepts of FL to the centralized data clustering algorithms. Then, we propose our dynamic GAN-based clustering algorithm in Section~\ref{sec:Dynamic_clustering}. Section~\ref{sec:experiments} describes the setting for the network and ML models, then the results from the experiments are described. Finally, we conclude our paper in Section~\ref{sec:conclusion}.

\section{Related Work}
\vspace{-1.5mm}
\label{sec:related_work}
The authors in \cite{zhao2018federated} have analyzed the non-IID data, which has degraded the performance of the ML model, e.g, convolutional neural network for classification tasks. In the paper, a small subset of client's data has improved the performance. However, the approach has shown a problem since the data transmission for the subset has leaked information about the training data. 

The algorithms in \cite{huang2019patient, briggs2020federated, ghosh2019robust} have tried to improve the performance of the ML models by applying clustering algorithms. The approach in \cite{huang2019patient} has compressed the data by an autoencoder and created clusters of the clients by the compressed data. Although the algorithm has improved ML models, it is subject to privacy breach since the data have been able to be restored by the compressed data and the autoencoder.

The authors in \cite{briggs2020federated} have created clusters by the similarities of gradients. This approach has shown a limitation due to that applying differential privacy (DP) has been challenging. In \cite{zhu2019deep, melis2019exploiting}, the authors have shown that the gradients have been able to reveal the original data, and DP has assured data privacy. DP has added noise to gradients, and the noises have been attenuated by averaging models. However, when applying DP to \cite{briggs2020federated}, the noises have been not able to be attenuated, which has been able to result in incorrect clusters. Next, the authors in \cite{ghosh2019robust} have clustered a local empirical risk minimizer by k-means. Although the algorithm in \cite{ghosh2019robust} avoids transmitting original data, it still creates fixed-size clusters.

The existing clustering algorithms have been non-dynamic, but the algorithm in \cite{mansour2020three} introduces the dynamic clustering. This algorithm has relocated clients to the best-fitting clusters by the performances of ML models, but there have been two drawbacks. First, depending on the quality of initial clusters, the time to converge to correct clusters has been able to vary. Second, the number of clusters has been still fixed. Thus, the clusters have not converged to the optimal number, because the fixed number of clusters has been able to be either an under-estimated or over-estimated value of the optimal number.

\section{Background}
\label{sec:background}
\vspace{-2mm}
\subsection{Federated Learning}
\label{sec:FL}
FL is a method to train a ML model when data is geographically distributed, as opposed to distributed in a datacenter \cite{mcmahan2016communication}, which consists of clients and a server. 
Stochastic gradient descent (SGD) can be used to train ML models in FL, and the training continues until convergence of the ML model. When the ML model is built in the server, the training round is as follows. First, the server selects a subset of clients and transmits the model to the subset. Second, each client trains the received model on local data and transmits the trained model to the server. Third, the server averages the trained models. 

\subsection{ClusterGAN}
\label{sec:clusterGAN}
ClusterGAN in \cite{mukherjee2019clustergan} creates clusters by three ML models that aim to compress data to a low dimension called a latent space. To be specific, the latent space consists of the Normal distribution and one-hot vector as follows:  
\begin{equation}\label{eq:CGAN_input}
\begin{split}
 z &= (z_{n},z_{c}),\\
  z_{n} &\sim \mathcal{N}(0,\sigma^{2} I_{d_{n}}),\\
  z_{c} &= e_{k},k \sim \mathcal{U} \{ 1,K \},
\end{split}
\end{equation}
  where $z$ is a variable in the latent space, $z_{n}$ is a noise of Normal distribution, $\mathcal{N}(0, \sigma^{2} I_{d_{n}})$ is the Normal distribution with mean 0 and variance $\sigma^{2} I_{d_{n}}$, $I_{d_{n}}$ is an identity matrix of size $d_{n}$, $z_{c}$ is a vector which stands for a cluster-ID, $e_{k}$ is a one-hot vector where all entries are zero except for one at $k$-th entry, and $\mathcal{U}( 1 , K )$ is a uniform random distribution with the lowest value 1 and the highest value $K$. In detail, the clusterGAN consists of the three ML models in Figure~\ref{fig:Structure_ClusGAN}. 

\begin{figure}[t]
  \begin{center}
        \includegraphics[width=0.3\textwidth]{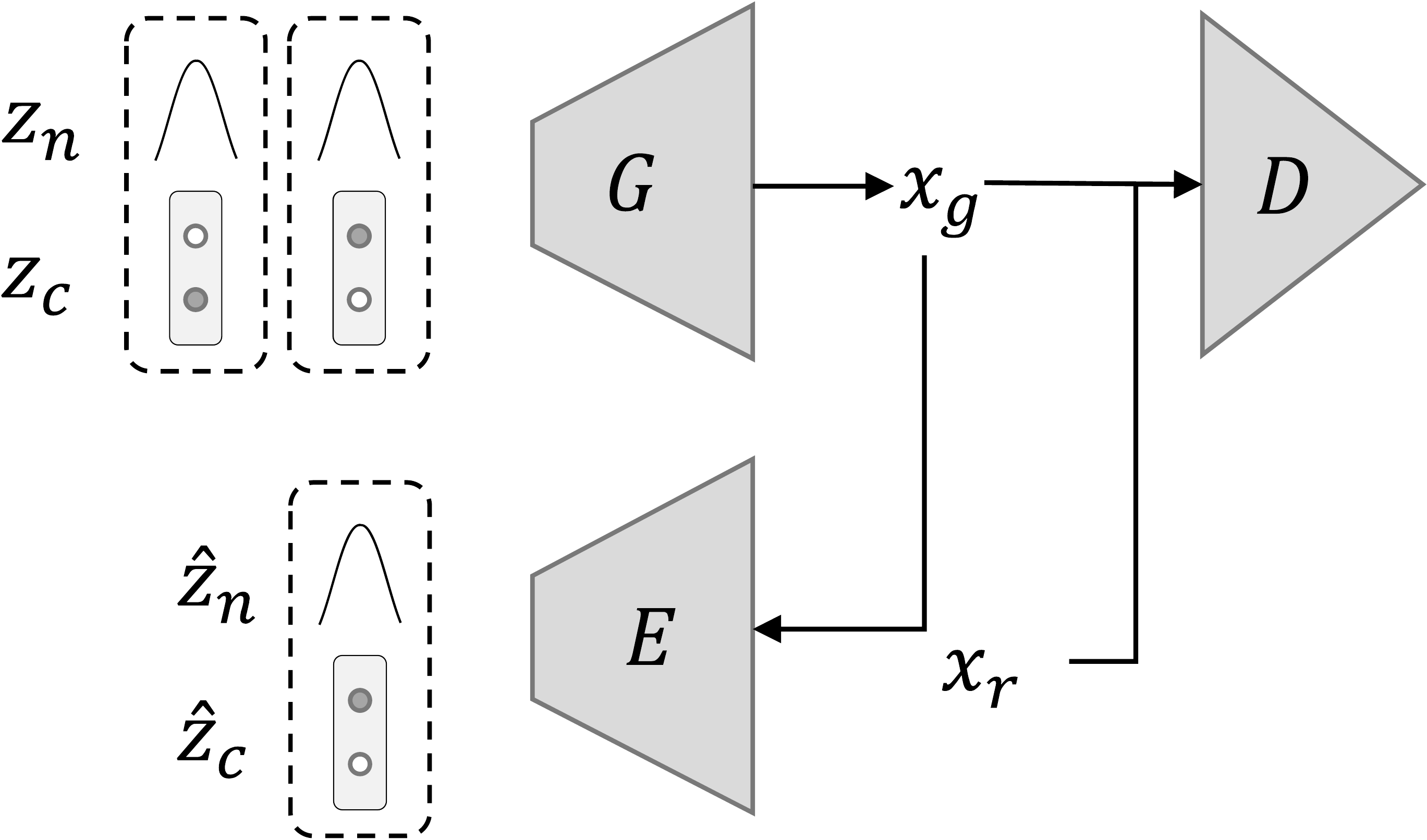}
  \end{center}
  \caption{The structure of the clusterGAN with the generator, encoder, and discriminator. The generator synthesizes data from a latent space, and the encoder maps the synthetic data into the latent space. The discriminator distinguishes between real data and synthetic data.}
  \label{fig:Structure_ClusGAN}
\end{figure}
 In Fig. 1, $G$, $E$, and $D$ are the models of the generator, encoder, and  discriminator. The models have weights which are parameterized by $\Theta_{G}$, $\Theta_{E}$ and $\Theta_{D}$, and  \textbf{$x_{g}$} and \textbf{$x_{r}$} are synthetic and real data. Let $\mathcal{Z}$ and $\mathcal{X}$ be the latent space and data space, and $\mathbb{R}$ be the set of real numbers. The functionality of the three models are as follows:
\begin{itemize}
    \item \textbf{Generator}: It maps the values in the latent space to the data space (${G}$: $\mathcal{Z} \to \mathcal{X}$);
    \item \textbf{Encoder}: It maps the values in the data space to the latent space (${E}$: $\mathcal{X} \to \mathcal{Z}$);
    \item \textbf{Discriminator}: It distinguishes whether the input data are synthetic data by the generator or real data (${D}$: $\mathcal{X} \to \mathbb{R}$).
\end{itemize}
 To train the three models, the loss function is as follows:
\begin{multline}
\min_{\Theta_{G}, \Theta_{E}} \max_{\Theta_{D}} \mathbb{E}_{x \sim P^{r}_{x}}q(D(x)) + \mathbb{E}_{z \sim P_{z}}q(1 - D(G(z))) \\ + \beta_{n}\mathbb{E}_{z \sim P_{z}} \vert \vert z_{n} - \mathcal{E}(\mathcal{G}(z_{n})) \vert \vert^{2}_{2} + \beta_{c}\mathbb{E}_{z \sim P_{z}}\mathcal{H}(z_{c},\mathcal{E}(\mathcal{G}(z_{c}))),
\label{eq:ClusterGAN_loss}
\end{multline}
where $\mathbb{E}$ is the expectation, $x$ is a real data, $P^{r}_{x}$ is the distribution of real data samples, $P_{z}$ is the distribution of noise in the latent space, $q(\cdot)$ is the quality function which is $\log(x)$ for vanilla GAN and $x$ for Wasserstein GAN \cite{arjovsky2017wasserstein},  $\beta_{n}$ and $\beta_{c}$ are hyperparameters, and $\mathcal{H}(\cdot)$ is the cross-entropy loss \cite{Goodfellow-et-al-2016}.  

\subsection{Hypothesis-based clustering (HypCluster)}
\label{sec:hypcluster}
HypCluster starts with $q$ hypothesis models, where $q$ is the number of clusters \cite{mansour2020three}. This algorithm calibrates clusters dynamically. In detail, a ML model is trained in each cluster, and each client runs the updated models and is relocated to the best-fitting cluster. This is detailed in Algorithm~\ref{alg:hyp}.
    \begin{algorithm}
    \SetKwInOut{Initialize}{Initialize}
        \caption{HYPCLUSTER \cite{mansour2020three}}\label{alg:hyp}
        \Initialize{Randomly sample $p$ clients among $P$ clients, train a model on them, and initialize each hypothesis model $h^{0}_{i}$ for all $i\in[q]$.}
        \For {$t=1$ to $T$}{ 
             Randomly sample $p$ clients\\
             Recompute $f^{t}$ for clients in $P$ by assigning each client to the cluster that has the lowest loss:    
            \begin{align} \label{eq:HypC_reshape} f^{t}(k) = \argmin_{i} \mathcal{L}_{\hat{\mathcal{D}_{k}}(h^{t-1}_{i})}. \end{align}\\
             Run $E$ steps of SGD for $h^{t-1}_{i}$ with data from clients $P \cap (f^{t})^{-1} (i)$ to minimize
 \begin{align}  \sum_{k:P \cap (f^{t})^{-1} (i)} m_{k} \mathcal{L}_{\hat{\mathcal{D}_{k}}}(h_{i}), \end{align}
  and obtain $h^{t}_{i}$
        }
         Compute $f^{T+1}$ by using $h^{T}_{1}, h^{T}_{2},...,h^{T}_{q}$ via \eqref{eq:HypC_reshape} and output it.
    \end{algorithm} 
    
In Algorithm~\ref{alg:hyp}, $h^{t}_{i}$ is the hypothesis model of cluster $i$ which has been updated $t$ times, $f$ is a function that maps a client to a cluster, $\mathcal{L}$ is a loss that occurs when a certain model is applied, $\hat{\mathcal{D}}_{k}$ is the empirical distribution on each client, $E$ is the epoch number, and $m_{k}$ is the number of data samples used for training at each client. Note that $q$ is the number of clusters by the hypothesis testing, and $P$ is the total number of clients. Therefore, it can be seen that the generalization decreases as $q$ increases. However, if $q = P$, i.e., the number of clusters is equal to the number of clients, each model becomes a local model of an individual client. Thus, the optimal $q$ less than $P$ must be found by testing several values of $q$.

\subsection{Divisive clustering}
\label{sec:divisive_clsutering}
Divisive clustering iteratively divides a cluster into two clusters until a stopping condition of clusters is satisfied \cite{savaresi2002cluster}. To divide the cluster, there are three concepts:
\begin{itemize}
    \item Divide all clusters into two clusters;
    \item Divide the cluster with the most data into two clusters;
    \item Divide the cluster with the highest variance by a distance metric (e.g., Euclidean distance) into two clusters.
\end{itemize}
The first concept does not consider the quality of clusters. The second can end up with balanced clusters, but it does not consider the quality of clusters. Thus, the third one is regarded as the most complex approach. 

\begin{figure}[!t]
  \begin{center}
        \includegraphics[width=0.7\linewidth ]{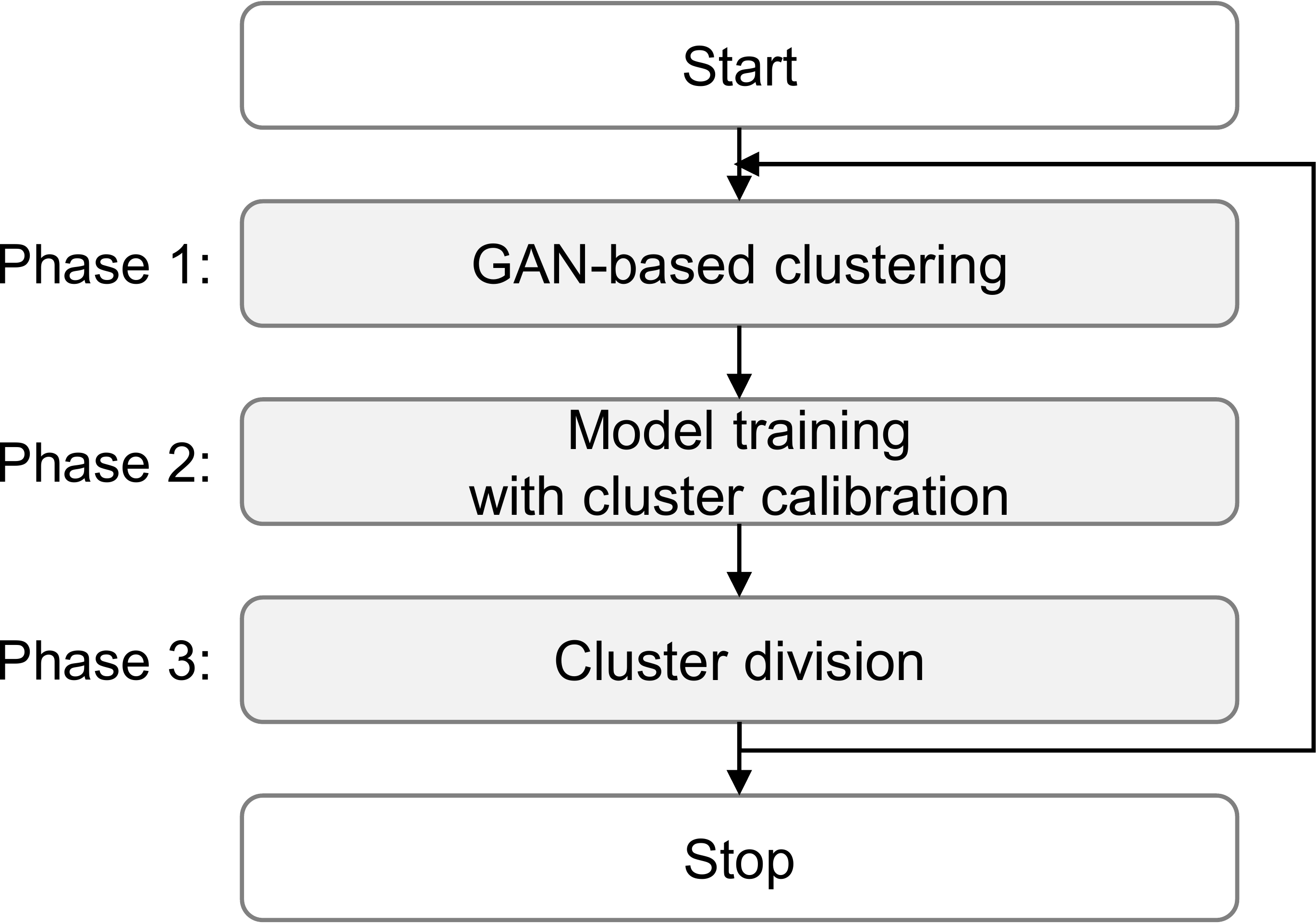}
  \end{center}
  \caption{Dynamic GAN-based clustering consists of three phases and executes the three phases sequentially. When the cluster division selects a cluster to divide, our algorithm performs Phase 1 and 2 on the selected cluster.}
  \label{fig:Dynamic_GAN_Based_Clustering}
\end{figure}

\section{Dynamic GAN-based Clustering}
\label{sec:Dynamic_clustering}
Our solution in Figure \ref{fig:Dynamic_GAN_Based_Clustering} consists of three phases:  GAN-based clustering, model training with cluster calibration, and cluster division.

 Phase 1 creates clusters by the GAN-based clustering which is a modification of clusterGAN in Section \ref{sec:clusterGAN} for FL. This phase preserves privacy and creates clusters (Algorithm~\ref{alg:Dynamic_GAN_Based_Clustering-Phase_1}).
    \begin{algorithm}[t]
    \SetKwInOut{Initialize}{Initialize}
        \caption{Phase 1: GAN-based clustering}
        \label{alg:Dynamic_GAN_Based_Clustering-Phase_1}
        \Initialize{$\mathcal{G}$, $\mathcal{D}$ and $\mathcal{E}$ by parameters $\Theta^{t}_{G}$, $\Theta^{t}_{D}$ and $\Theta^{t}_{E}$\label{line:phase1_start}} 
        \For {ClusterGAN round t = 1,...,r}{
                 $m$ $\leftarrow$ $\max{(\vert C\vert \cdot r,1)}$
                 $S_{t}$ $\leftarrow$ (sample $m$ clients in C)
                    \For {each client $c \in$ $S_{t}$ in parallel}{
                     Initialize the GAN at $c$ with $\Theta^{t}_{G}, \Theta^{t}_{D}, \text{ and } \Theta^{t}_{E}$ \\
                     Train the GAN $E_{1}$ times at $c$ using SGD on local data to obtain $\Theta^{t+1}_{G}, \Theta^{t+1}_{D}, \text{ and } \Theta^{t+1}_{E}$\\
                     Send $\Theta^{t+1}_{G}, \Theta^{t+1}_{D}, \text{ and } \Theta^{t+1}_{E}$ to the server \label{line:phase1_train}
                    }
                 $\Theta^{t+1}_{D}$ $\leftarrow$ $\sum_{c=1}^{m}{\frac{n_{c}} {n}}\Theta^{t+1}_{D_{c}}$ \\
                 $\Theta^{t+1}_{G}$ $\leftarrow$ $\sum_{c=1}^{m}{\frac{n_{c}} {n}}\Theta^{t+1}_{G_{c}}$ \\
                 $\Theta^{t+1}_{E}$ $\leftarrow$ $\sum_{c=1}^{m}{\frac{n_{c}} {n}} \Theta^{t+1}_{E_{c}}$
        }
        
        \For {each client $c \in C$ in parallel}{
         Initialize the encoder at $c$ with $\Theta^{r}_{E}$\\
         Infer cluster-IDs of all data samples at $c$\\
         $id^{c} \leftarrow$ the major cluster-ID at $c$ \label{line:phase1_infer}
        }  \label{line:phase1_end} 
    \end{algorithm}
    
    In Algorithm~\ref{alg:Dynamic_GAN_Based_Clustering-Phase_1}, $C$ is the set of all clients, $r$ is the ratio to sample clients, $m$ is the number of clients to sample, $\vert \cdot \vert$ is the cardinality of a set, and $S_{t}$ is the set of sampled clients.
    
Phase 2 trains ML models and calibrates clusters by HypCluster in Section \ref{sec:hypcluster}. Cluster calibration updates the fixed clusters by modifying the clusters to be performance-focused clusters as described in Section \ref{sec:hypcluster} (Algorithm~\ref{alg:Dynamic_GAN_Based_Clustering-Phase_2}).

    \begin{algorithm}[!t]

    \SetKwInOut{Initialize}{Initialize}
        \caption{Phase 2: model training with cluster calibration}
        \label{alg:Dynamic_GAN_Based_Clustering-Phase_2}

 \Initialize{$k$ models for a use case with     $w^{k}_{t+1}$} \label{line:phase2_start}
        \For {HypCluster round t' = 1,...,h}{
            \For {ML model training round t'' = 1,...,l}{
                $m'$ $\leftarrow$ $\max{(\vert C \vert \cdot r',1)}$\\
                $S'_{t}$ $\leftarrow$ sample $m'$ clients in C

                \For {each client $c \in$ $S'_{t}$ in parallel}{
                     Define the weight of the ML model for the cluster $id^{c}$ as $w^{id^c}_{t}$\\
                     Initialize the ML model at $c$ with $w^{id^c}_{t}$\\
                     Train the ML model $E_{2}$ times at $c$ using SGD and obtain the updated model $w^{id^c}_{t+1}$\\
                     Send  $w^{id^c}_{t+1}$ to the server \label{line:phase2_train}
                    }
                    
                \For {each client $c \in$ $S'_{t}$ }{
                     $w^{id^c}_{t+1}$ $\leftarrow$ $\sum_{c=1}^{C}{n_{c} \over n} $  ${w^{id}}^{c}_{t+1}$
                }
            }
              $W_{t}$ $\leftarrow$ $w^{id}_{t}$ $id \in$ \{1,...,k\}\\
            \For {each client $c \in \{1,...,C\}$ in parallel}{
                 Initialize all ML models at $c$ with $W_{t}$ \\
                 Run all models using local data at $c$\\
                Identify the cluster-ID of the model with the lowest loss and set the cluster-ID as $id^{c}$\\
                Send the $id^{c}$ to the server \label{line:phase2_calibration}
            }
        }
        \label{line:phase2_end} 
        
    \end{algorithm}

    \setlength{\textfloatsep}{1\baselineskip}
    \begin{algorithm}[!t]
    \SetKwInOut{Initialize}{Initialize}
        \caption{Phase 3: cluster division}
        \label{alg:Dynamic_GAN_Based_Clustering-Phase_3} 

        \Initialize{Set the initial number of cluster $k$ and the clients $C$ for the whole algorithm.}
        \For {Divisive clustering round $i$ = 1,...,I}{
             Perform Phase 1 and Phase 2\\
            \For {each client $c \in \{1,...,C\}$ in parallel}{
                 Initialize the ML model at $c$ with $w^{id^c}_{t}$\\
                 Run the model at $c$ to obtain the loss $loss^{id^c}$ \\
                 Send $loss^{id^c}$ to the server 
            }

            Calculate averages and variances of $loss^{id}$ for each $id \in \{1,...,k\}$ to obtain the list of averages $avg\_losses$  and variance  $var\_losses$ \label{line:phase3_start}\\
            
             Select a cluster $sel\_id$ to divide by applying the priority on  $avg\_losses$ and $var\_losses$ \label{line:phase3_prioritize}\\ 
             
             Set $C$ as the clients in cluster $sel\_id$\\
             Set $k$ as the new number of clusters $n$ (default: $n=2$) 
            }  \label{line:phase3_end}

    \end{algorithm}
In Algorithm~\ref{alg:Dynamic_GAN_Based_Clustering-Phase_2}, $r'$ is the ratio to sample clients, and $m'$ is the number of clients to sample. Note that our algorithm examines all models on the clients, and each client is relocated to the best-fitting cluster on line \ref{line:phase2_calibration} in Algorithm \ref{alg:Dynamic_GAN_Based_Clustering-Phase_2}.
    
Phase 3 modifies the divisive clustering in Section \ref{sec:divisive_clsutering} by using the mean and the variance of model performance. This phase selects and splits a cluster (Algorithm~\ref{alg:Dynamic_GAN_Based_Clustering-Phase_3}).

Note that a priority is applied to select a cluster on line~\ref{line:phase3_prioritize} in Algorithm \ref{alg:Dynamic_GAN_Based_Clustering-Phase_3}. The first priority is the cluster with the highest variance above a predefined variance threshold. This is because the ML model can show different performance for different data characteristics in a cluster. The second is the cluster with the highest mean above a predefined mean threshold. This is selected since the model can show low performance for different characteristics in a cluster. After Phase 3, our algorithm stops by two conditions: first, all clusters show lower variance and mean than the two predefined thresholds. Second, all divisive clustering rounds are finished.   

In Section \ref{sec:experiments}, we discuss more details about the execution of our proposed algorithm, including the number of rounds and the time scale of each phase.

\section{Experimental Setting and Results}
\label{sec:experiments}
\subsection{Experimental Setup}
We use three public datasets with class information and one private dataset without class information. Note that the data-type of all datasets is time series since our main goal, i.e., resource management by handover prediction, is time series forecasting. 
The details of the datasets are:   
\begin{table}[!t]
    \begin{center}
    \caption{Time series datasets}
    \label{tab:dataset}
\begin{tabular}{@{}cccc@{}}
\toprule
\textbf{Dataset name} & \textbf{No. of samples} & \textbf{Len. of sample} & \textbf{No. of class} \\ \midrule
Italy        & 1,096  & 24    & 2       \\
Pendigit     & 10,992 & 16    & 10      \\
Melbourne    & 3,633  & 24    & 10      \\
Handover & 149    & 1,392 & Unknown \\ \bottomrule
\end{tabular}
    \end{center}
    \vspace{-3mm}
\end{table}

\begin{itemize}
    \item \textbf{Handover:} The number of handovers across 149 smaller geographical areas in a metropolitan city were hourly counted for 58 days. This data can reveal the movement of the population in a city. Thus, it is desirable to preserve data privacy of user mobility data. However, the data can also benefit resource management and therefore it is of interest to use the data for forecasting;
    
    \item \textbf{Italy power demand:} This is a dataset that recorded the twelve-monthly power demand in 1997 \cite{keogh2006intelligent}; 
    
    \item \textbf{Pendigit:} The coordinates of moving pens were recorded as time series when writing digits \cite{Dua:2019}; 
    
    \item \textbf{Melbourne pedestrian:} In Melbourne, the number of pedestrians was counted by pedestrian counters \cite{mbr}. 
    
\end{itemize}

We normalize each dataset in Table~\ref{tab:dataset} depending on its own characteristics. To be specific, for Pendigit and Italy power demand, the range of each dataset is set to $[0, 1]$ by min-max scaler. The reasons are as follows: the range of Pendigit was normalized to $[0,100]$ in \cite{Dua:2019}. Regarding the Italy power demand, we aim to distinguish different amounts of power demands. On the other hand, for the Melbourne pedestrian and Handover, the min-max scaler is applied to each time series since we aim to distinguish the different shapes of time series.

The three public datasets are distributed to 30 clients. We divide each dataset into the training, overwriting, and test data, i.e., 70\%, 20\%, and 10\% of each dataset. Each client has the training data and test data of a single class to simulate a non-IID environment. The overwriting data are used to simulate a dynamic environment in Phase 2. Specifically, in the middle of Phase 2, one client is randomly selected, and its all data are overwritten by the data of a random class. 

For Handover, we simulate the entire city. Thus, we create 149 clients with the time series from time step 0 to 1,344 as training data. The remaining 48 steps are used to test long short term memory (LSTM) models. For Phase 1, the time series of each week is a sample. To train the LSTM models, we augment the training data with the shift window technique by selecting 19 time steps and shifting one time step. Note that the dynamic environment is not simulated for two reasons: we aim to simulate a city with 149 cells, and the dynamic environment cannot be controlled without the class information. 

In Phase 1, the model of GAN-based clustering follows the architecture for time series in \cite{mukherjee2019clustergan} except for the Handover dataset. For Handover, we use 512 neurons in each layer and 80 for the size of $z_{n}$. Note that the number of clusters is the number of classes for the public datasets and two for the private dataset. This is because we aim to divide clusters iteratively when the number of classes is unknown. 
\begin{table}[!t]
  \begin{center}
    \caption{LSTM model}
    \label{tab:lstm_model}
\begin{tabular}{@{}lcccc@{}}
\toprule
\textbf{}                  & \textbf{Italy} & \textbf{Pendigit} & \textbf{Melbourne} & \textbf{Handover} \\ \midrule
\textbf{LSTM-neurons}      & 8              & 8                 & 8                  & 8              \\
\textbf{Batch size}        & 7              & 7                 & 7                  & 2              \\
\textbf{Learning rate}     & 0.001          & 0.001             & 0.001              & 0.001          \\
\textbf{Step size rule}         & RMSProp        & RMSProp           & RMSProp            & RMSProp        \\
\textbf{L2 regularization} & 0.0005         & 0.0005            & 0.0005             & 0.0005         \\ \bottomrule
\end{tabular}
  \end{center}\vspace{-3mm}
\end{table}

In Phase 2, the structure of LSTM model is in Table \ref{tab:lstm_model}. The training data are divided into the input and the desired output of LSTM. In our experiment for the public datasets, the time series from the beginning to the 70\% of the length of each sample is the input, and the rest becomes the desired output. Regarding Handover, we aim to simulate the harsh prediction condition for the real network, where 60\% of 19 time steps is the input, and the rest becomes the desired output of LSTM. Also, for all datasets, if the length of each sample is not an integer, we round down the number.

In Phase 3, the number of the divisive rounds is one for the public datasets and ten for the private dataset. Note that Phase 3 for the public datasets divides a cluster once since the number of clusters by Phase 1 should be close to the number of classes. For the private dataset, the cluster division occurs ten times since we aim to start from the under-estimated number of clusters and vary the number of clusters over time. Regarding the threshold of the variance and mean, we select 1.0e-6 for the variance and 1.0e-2 for the mean.

In our simulation, the GAN-based clustering model in Phase 1 is trained for 50,000 rounds. For Phase 2, LSTM models are trained for 100 global rounds with two steps of local training. Thus, 200 steps of the LSTM training are carried out. Each round is triggered every hour since we aim to train LSTM models with a new sample. During Phase 2, cluster calibration is triggered at 40th and 80th rounds. After Phase 2, Phase 3 selects a cluster to divide. Then, our algorithm executes Phase 1 but reduces the number of rounds for Phase 1 to 25,000.

Regarding the time scale in a wireless network scenario (i.e. Handover), the first execution of Phase 1 and Phase 2 takes 15 hours and 100 hours. Thus, after executing our algorithm for 115 hours, the first execution of Phase 3 is triggered. 

To numerically verify our algorithm, the result of our algorithm was compared to a baseline algorithm using a feature extraction algorithm in \cite{hyndman2015large} and agglomerative clustering in \cite{rokach2005clustering}. 
    Specifically, the chosen feature extraction algorithm extracts the features on each client, and the extracted features are sent to the server. The clustering algorithm executes using these features. For the baseline, the dynamic environment is not simulated since the baseline cannot handle the environment. 
    
    \subsection{Evaluation metrics}
    We show the numerical performance of our clustering algorithms and LSTM models. For the clustering algorithms, purity \cite{cambridge2009introduction} is calculated for the public datasets as follows: 
    \begin{align}
    \label{eqn:purity} \text{Purity} = {1 \over N} \sum^{k}_{i=1} \max_{j}|c_{i}\cap t_{j} |,
    \end{align}
    where $N$ is the number of data samples, $k$ is the number of clusters, $c_{i}$ is one of the $k$ clusters, and $t_{j}$ is the number of samples when class $j$ is the majority in the cluster $c_{i}$.
    
    Regarding the performance of LSTM models, the loss of LSTM models is shown since the accurate clusters enable the lower loss of LSTM models. The loss of LSTM models is measured by the mean squared error (MSE) as follows:
  \begin{align}
      \mathcal{L}(\textbf{y}, \hat{ \textbf{y}}) = {1 \over m} \sum_{i=1}^{m}(y_{i}- \hat{y_{i}})^{2},
      \label{eq:MSE} 
  \end{align}
  where $y_{i}$ and $ \hat{ {y}_{i}}$ are the output and desired output of LSTM models, and $m$ is the number of samples.
\subsection{Results}
\label{sec:results_and_discussion}
In the case of the datasets with the class information, the baseline and Phase 1 create a cluster-ID for each sample. Hence, the purities for the three public datasets are calculated and compared with the purities of the baseline. Also, the performances of LSTM models of all datasets were compared. 
\subsubsection{Result of Phase 1}
\begin{table}[!t]
      \begin{center}
        \caption{Purity of Phase 1 (unit: \%)}
        \label{tab:purity_phase1}
\begin{tabular}{@{}ccccc@{}}
\toprule
\multirow{2}{*}{\textbf{}} &
  \multirow{2}{*}{\textbf{Baseline}} &
  \multicolumn{2}{c}{\textbf{Phase 1}} &
  \multirow{2}{*}{\textbf{\begin{tabular}[c]{@{}c@{}}Improvement\\      (Average)\end{tabular}}} \\ \cmidrule(lr){3-4}
                   &       & \textbf{Average} & \textbf{Std Dev} &       \\ \midrule
\textbf{Italy}     & 65.02 & 76.60            & 6.08             & 11.58 \\
\textbf{Pendigit}  & 43.26 & 69.30            & 3.44             & 26.04 \\
\textbf{Melbourne} & 58.28 & 59.22            & 3.03             & 0.94  \\  \bottomrule
\end{tabular}
      \end{center}\vspace{-3mm}
\end{table}
The baseline and Phase 1 of our algorithm are repeated 5 times for statistical accuracy in Table~\ref{tab:purity_phase1}. Note that Handover is not included since the purity cannot be calculated without class information in Eq. \eqref{eqn:purity}. In Table~\ref{tab:purity_phase1}, the baseline does not have the standard deviation since the baseline always creates the same clusters. However, the clusters by Phase 1 can vary because of the clusterGAN. Thus, the standard deviations are in Table \ref{tab:purity_phase1}. From the table, we conclude that our Phase 1 outperforms the baseline since our algorithm shows higher purity than the baseline. 

\subsubsection{Result of Phase 2}

  \begin{figure} \vspace{-3mm}
    \centering
  \subfloat[\label{fig:italy_lstm}]{%
       \includegraphics[width=0.5\linewidth]{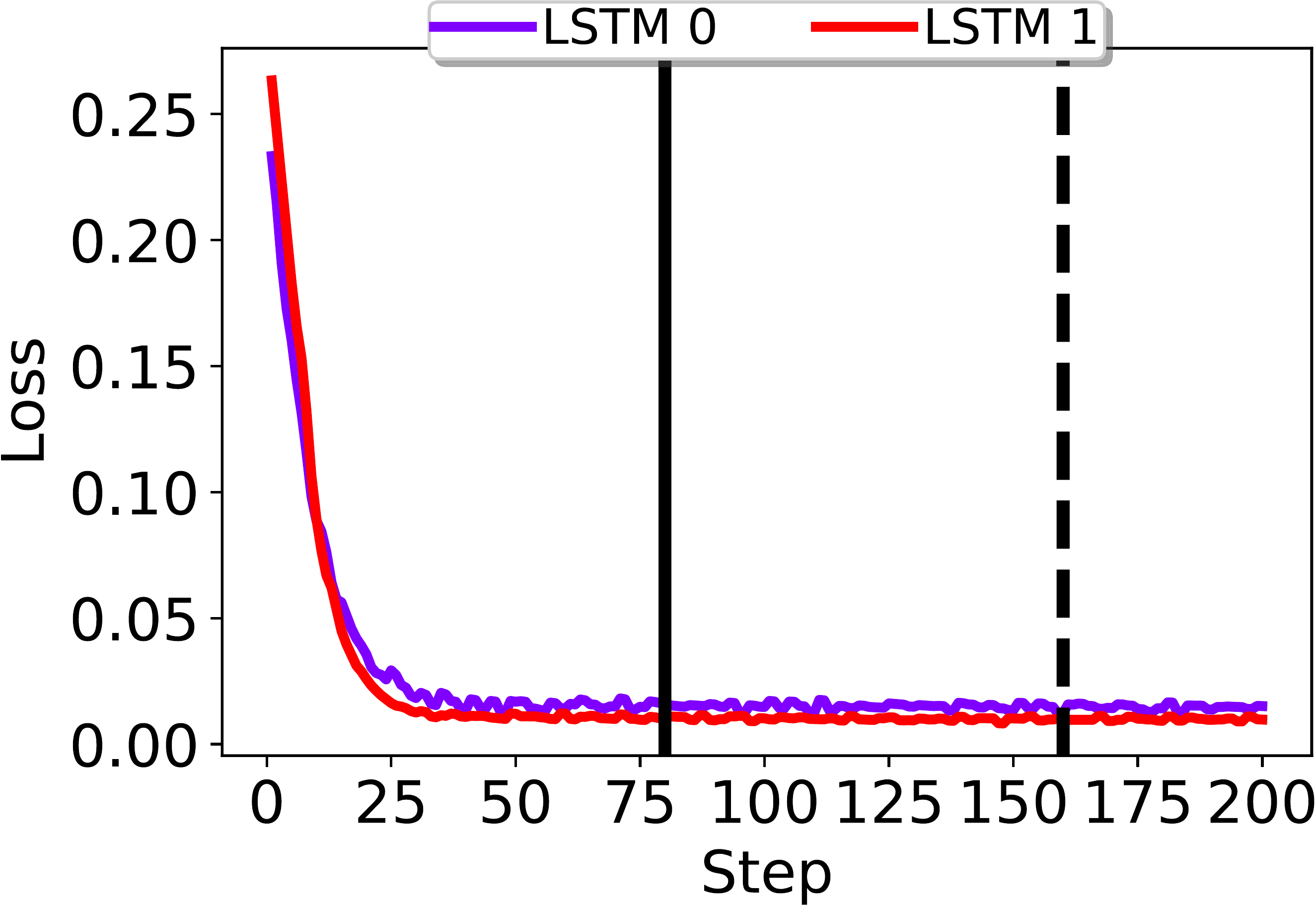}}
    \hfill
  \subfloat[\label{fig:pendigit_lstm}]{%
        \includegraphics[width=0.5\linewidth]{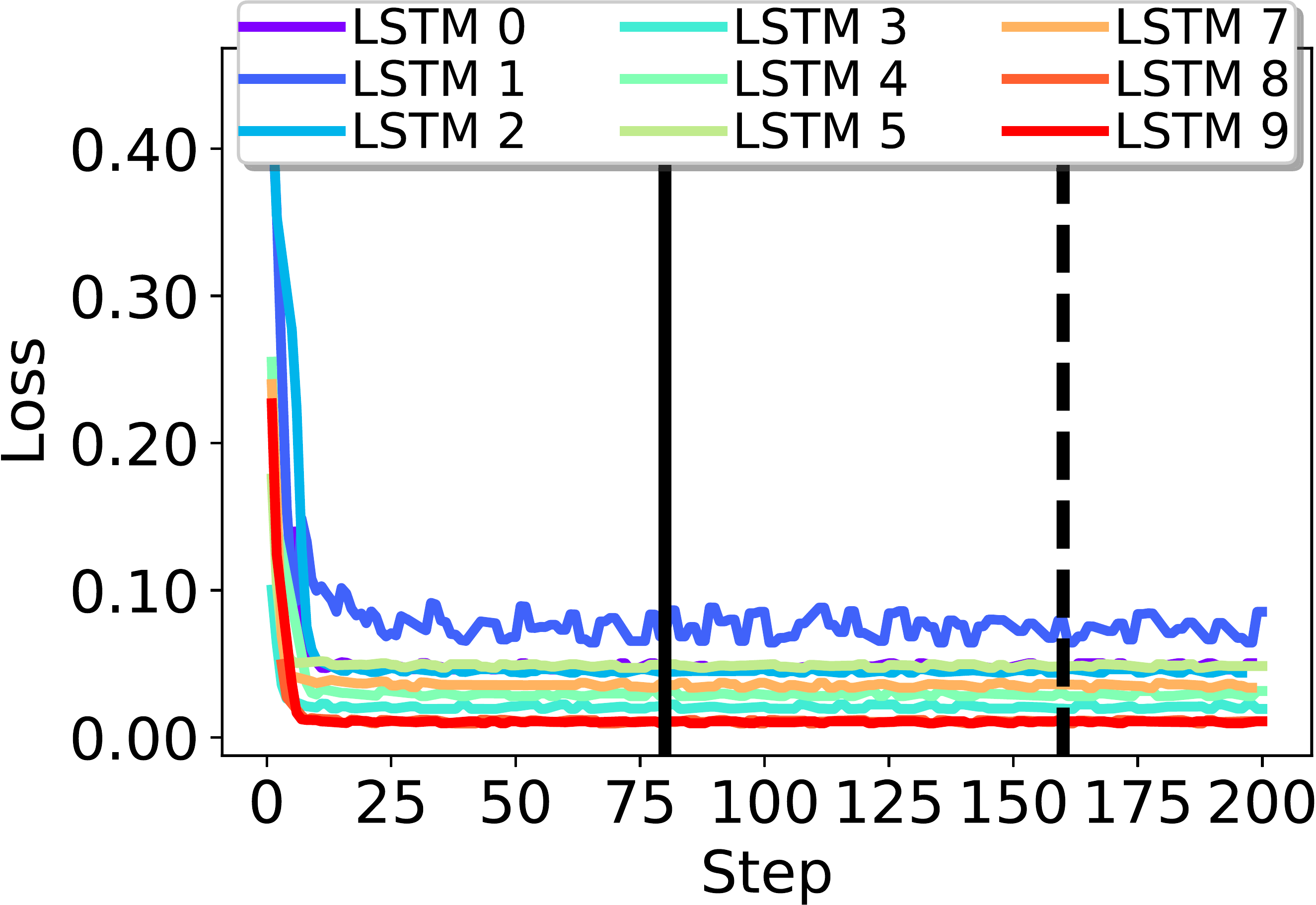}}
    \\\vspace{-3mm}
  \subfloat[\label{fig:mbr_lstm}]{%
        \includegraphics[width=0.5\linewidth]{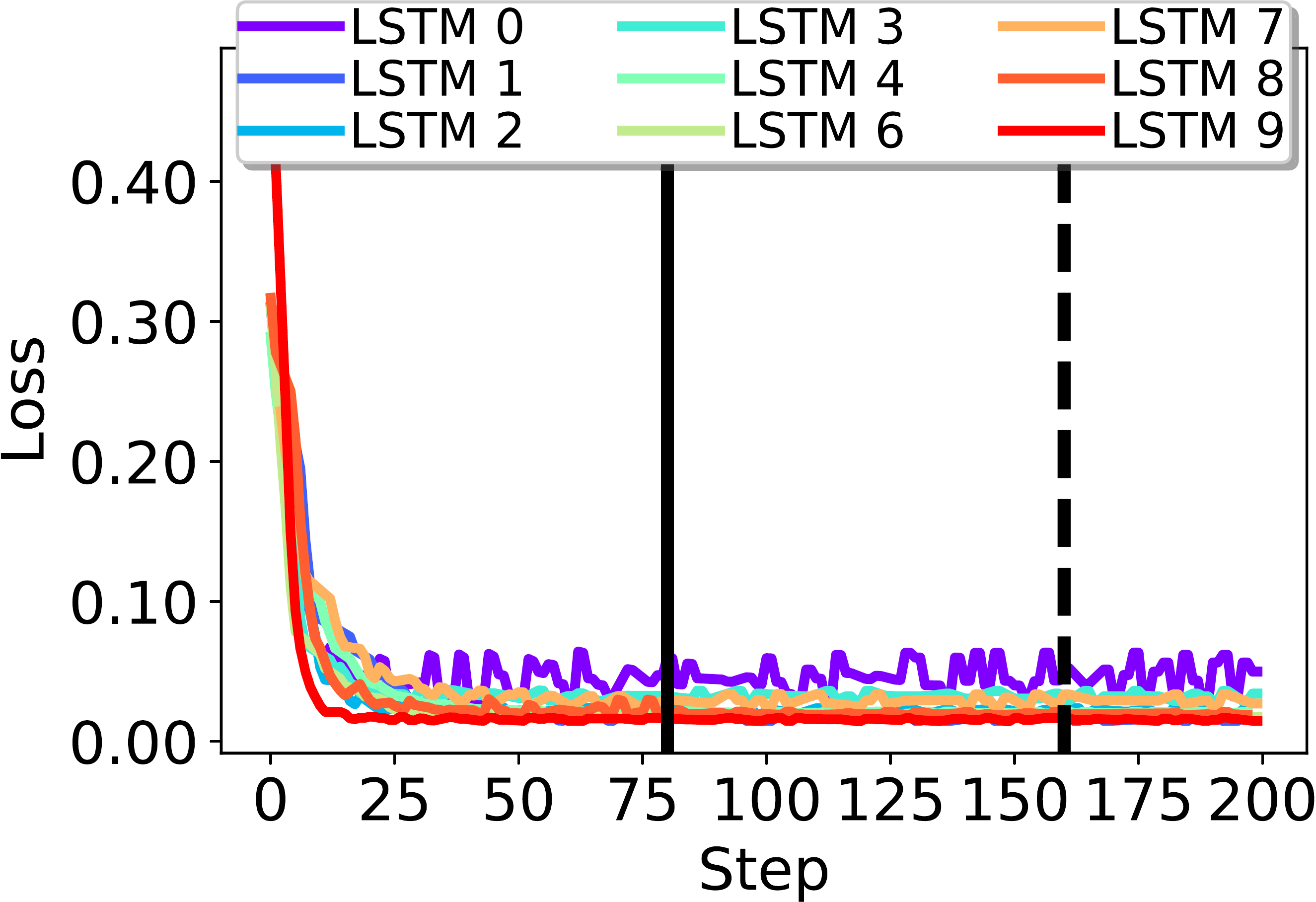}}
    \hfill
  \subfloat[\label{fig:mc_lstm}]{%
        \includegraphics[width=0.5\linewidth]{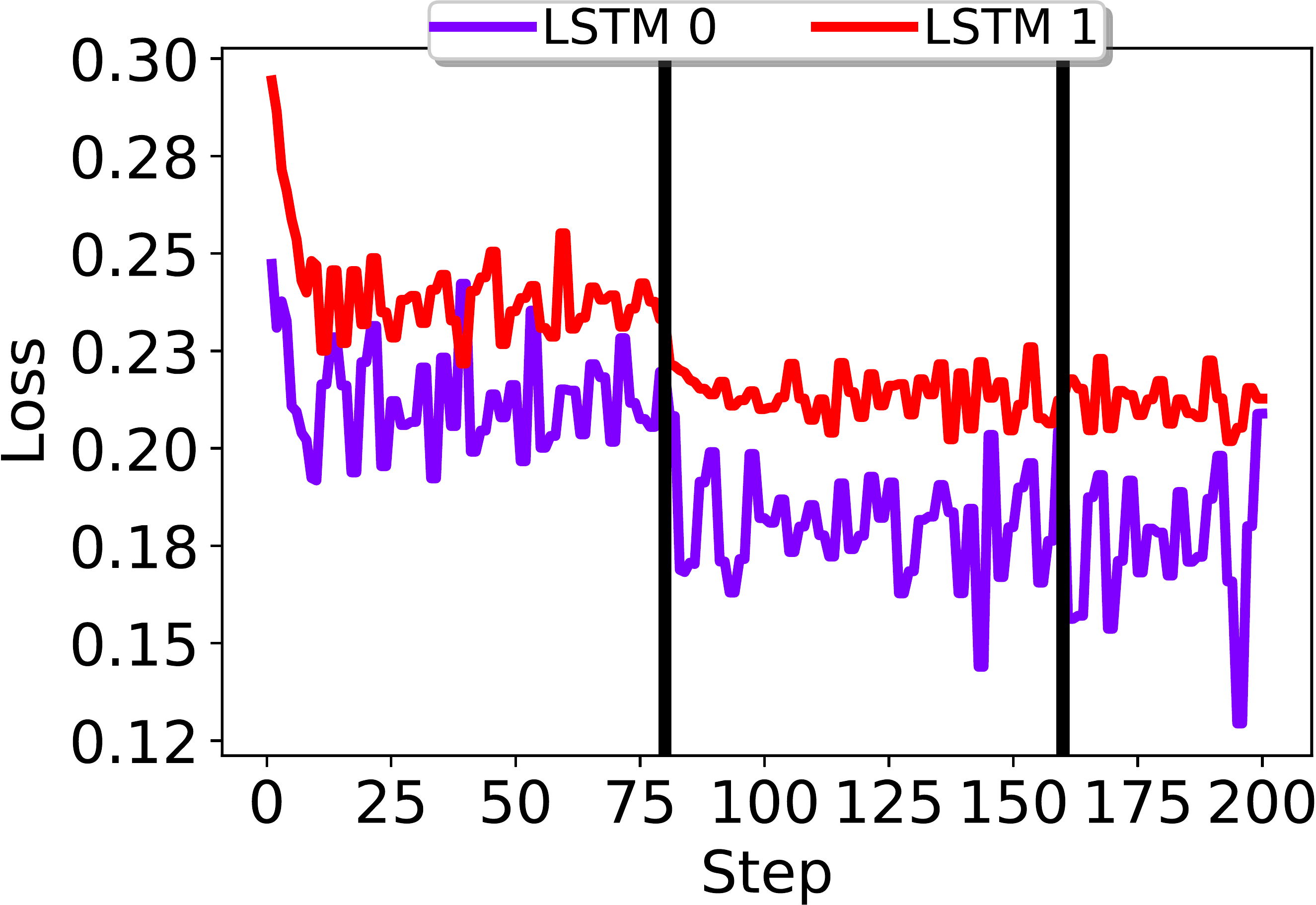}}
  \caption{LSTM training loss in the first divisive round of (a) Italy power demand, (b) Pendigit, (c) Melbourne pedestrian, and (d) Handover. The solid and dashed lines mean the cluster calibration in the static and dynamic environments.}
  \label{fig:lstm_training_loss}
\end{figure}

The cluster calibration is simulated in static and dynamic environments for the three public datasets, while Handover is simulated in a static environment. The baseline is not plotted due to two reasons. First, although we aim to show the benefits from the calibration, the calibration is not implemented by the baseline~\cite{hyndman2015large}. Second, the final performance of LSTM models is in Table~\ref{tab:purity_phase3}. Figure~\ref{fig:lstm_training_loss} shows the losses in the first divisive round. The vertical solid or dashed lines indicate the cluster calibration in the static or dynamic environment. As a result, we observe that the cluster calibration decreases the losses or prevents the possible fluctuation of losses by the dynamic environments.

\begin{table}[!t]

\vspace{-4mm}
      \begin{center}
        \caption{LSTM MSE of final results}
        \label{tab:purity_phase3}

\begin{tabular}{@{}cccccc@{}}
\toprule
\multirow{2}{*}{} &
  \multicolumn{2}{c}{\textbf{Baseline}} &
  \multicolumn{2}{c}{\textbf{Dynamic clustering}} &
  \multirow{2}{*}{\textbf{\begin{tabular}[c]{@{}c@{}}Improvement\\      (Average)\end{tabular}}} \\ \cmidrule(lr){2-5}
                      & \textbf{Average} & \textbf{Std Dev} & \textbf{Average} & \textbf{Std Dev} &         \\ \midrule
\textbf{Italy}        & 0.0123           & 0.000            & 0.0088           & 0.0003           & 28.68\% \\
\textbf{Pendigit}     & 0.0550           & 0.001            & 0.0305           & 0.0031           & 44.54\% \\
\textbf{Melbourne}    & 0.0320           & 0.001            & 0.0238           & 0.0010            & 25.61\% \\
\textbf{Handover} & 0.1942           & 0.002            & 0.1100           & 0.0021           & 43.36\% \\\bottomrule
\end{tabular}
      \end{center}\vspace{-3mm}
\end{table}
\subsubsection{Result of Phase 3}
\label{sec:result_phase3}
After the whole algorithm, LSTM models are tested. The loss of each LSTM model is weighted by the number of clients in each cluster and averaged. From the five repetitions of the simulations for each dataset, the results are summarized in Table~\ref{tab:purity_phase3}, and we observe that our algorithm outperforms the baseline for all datasets. 

However, the experiment also showed two drawbacks of our algorithm: first, Phase 1 consumes more time than the baseline since Phase 1 requires model training. Second, the cluster division can build clusters with a single client. 

\section{Conclusion and Future Work}
\label{sec:conclusion}

This paper introduced dynamic GAN-based clustering in FL to improve the accuracy of time series forecasting. The improvement is achieved by a novel policy to dynamically update the clusters over time while preserving privacy. From extensive simulations, our algorithm showed numerical improvement for time series forecasting from 29\% to 45\% compared to the baseline on four real-world datasets. 

For future works, we could leverage local computation power to speed up the convergence of the models. Next, we can consider a further policy of merging dynamically generated clusters to decrease the number of clusters.



\bibliography{biblio.bib}

\end{document}